%% file: sample-1col.tex
\documentclass[
]{ceurart}

\usepackage{listings}
\newcommand{\ours}{\text{SP-Guard}\xspace}
\usepackage{wrapfig}
\usepackage{algorithm}
\usepackage{algpseudocode}
\usepackage[capitalise]{cleveref}
\usepackage{subcaption}
\usepackage{mathtools}
\usepackage{enumitem}
\usepackage{amssymb}
\usepackage{xurl}
\usepackage{hyperref} 
\usepackage{float}
\usepackage{tabularx}
\usepackage[table]{xcolor}
\usepackage{graphicx}
\usepackage{caption}
\usepackage{chngcntr}
\definecolor{mygreen}{RGB}{0,176,80}
\definecolor{myred}{RGB}{255,0,0}
\definecolor{myblue}{RGB}{0,112,192}
\usepackage{xspace}
\newcommand{\ie}{\textit{i.e.}\xspace}
\usepackage{booktabs}
\usepackage{setspace} 
\usepackage[rightcaption]{sidecap}

\begin{document}

\copyrightyear{2025}
\copyrightclause{Copyright for this paper by its authors.
  Use permitted under Creative Commons License Attribution 4.0
  International (CC BY 4.0).}

\conference{TRUST-AI: The European Workshop on Trustworthy AI. Organized as part of the European Conference of Artificial Intelligence - ECAI 2025. October 2025, Bologna, Italy.}

\title{SP-Guard: Selective Prompt-adaptive Guidance for Safe Text-to-Image Generation}

\author[1]{Sumin Yu}[%
orcid=0009-0008-9752-0112,
email=ysmsoomin@snu.ac.kr,
url=https://sumin-yu.github.io,
]
\address[1]{Department of Electrical and Computer Engineering, Seoul National University, Seoul, South Korea}

\author[1,2]{Taesup Moon}[%
orcid=0000-0002-9257-6503    ,
email=tsmoon@snu.ac.kr,
url=https://mindlab-snu.notion.site/taesup-moon,
]
\cormark[1]
\address[2]{IPAI / ASRI / INMC, Seoul National University, Seoul, South Korea}

\cortext[1]{Corresponding author.}

\input{sec-ecai/0_abstract}


\maketitle

\input{sec-ecai/0_abstract}
\input{sec-ecai/1_intro}
\input{sec-ecai/3_method}
\input{sec-ecai/4_experiments}
\input{sec-ecai/5_discussion}
\begin{acknowledgments}
This work was supported in part by the National Research Foundation of Korea (NRF) grant [No.2021R1A2C2007884] and by Institute of Information \& communications Technology Planning \& Evaluation (IITP) grants [RS-2021-II211343, RS-2021-II212068, RS-2022-II220113, RS-2022-II220959] funded by the Korean government (MSIT).
\end{acknowledgments}

\section*{Declaration on Generative AI}
 During the preparation of this work, the author used Grammarly in order to check grammar and spelling. After using this tool, the author reviewed and edited the content as needed and takes full responsibility for the publication’s content.

\bibliography{sample-ceur}

\end{document}

%% file: sec-ecai/0_abstract.tex
\begin{abstract}
While diffusion-based T2I models have achieved remarkable image generation quality, they also enable easy creation of harmful content, raising social concerns and highlighting the need for safer generation.
Existing inference-time guiding methods lack both adaptivity—adjusting guidance strength based on the prompt—and selectivity—targeting only unsafe regions of the image.
Our method, SP-Guard, addresses these limitations by estimating prompt harmfulness and applying a selective guidance mask to guide only unsafe areas. 
Experiments show that SP-Guard generates safer images than existing methods while minimizing unintended content alteration.
Beyond improving safety, our findings highlight the importance of transparency and controllability in image generation.

\noindent
\textcolor{red}{\footnotesize \textbf{WARNING}: This paper contains AI-generated images that may be offensive.
Sensitive contents are masked.}
\end{abstract}

%% file: sec-ecai/1_intro.tex
\section{Introduction}
\label{sec:intro}
The rapid advancements in text-to-image (T2I) diffusion models \cite{T2Imodel:SD, 
T2Imodel:saharia2022photorealistic} have enabled the generation of high-quality images based on textual inputs. However, the extensive training data often contain unsafe content and inherent biases \cite{T2Imodel:SD, LAION:schuhmann2022laion}, posing significant risks of generating unexpected unsafe images \cite{birhane2021multimodal}. There are also concerns about malicious users exploiting model vulnerabilities to create harmful images by generating attacking prompts \cite{ring-a-bell:tsai2023ring, mma-diffusion:yang2024mma, guardT2I:yang2024guardt2i}.

\begin{SCfigure}[1.15][t]
    \includegraphics[width=0.45\linewidth]{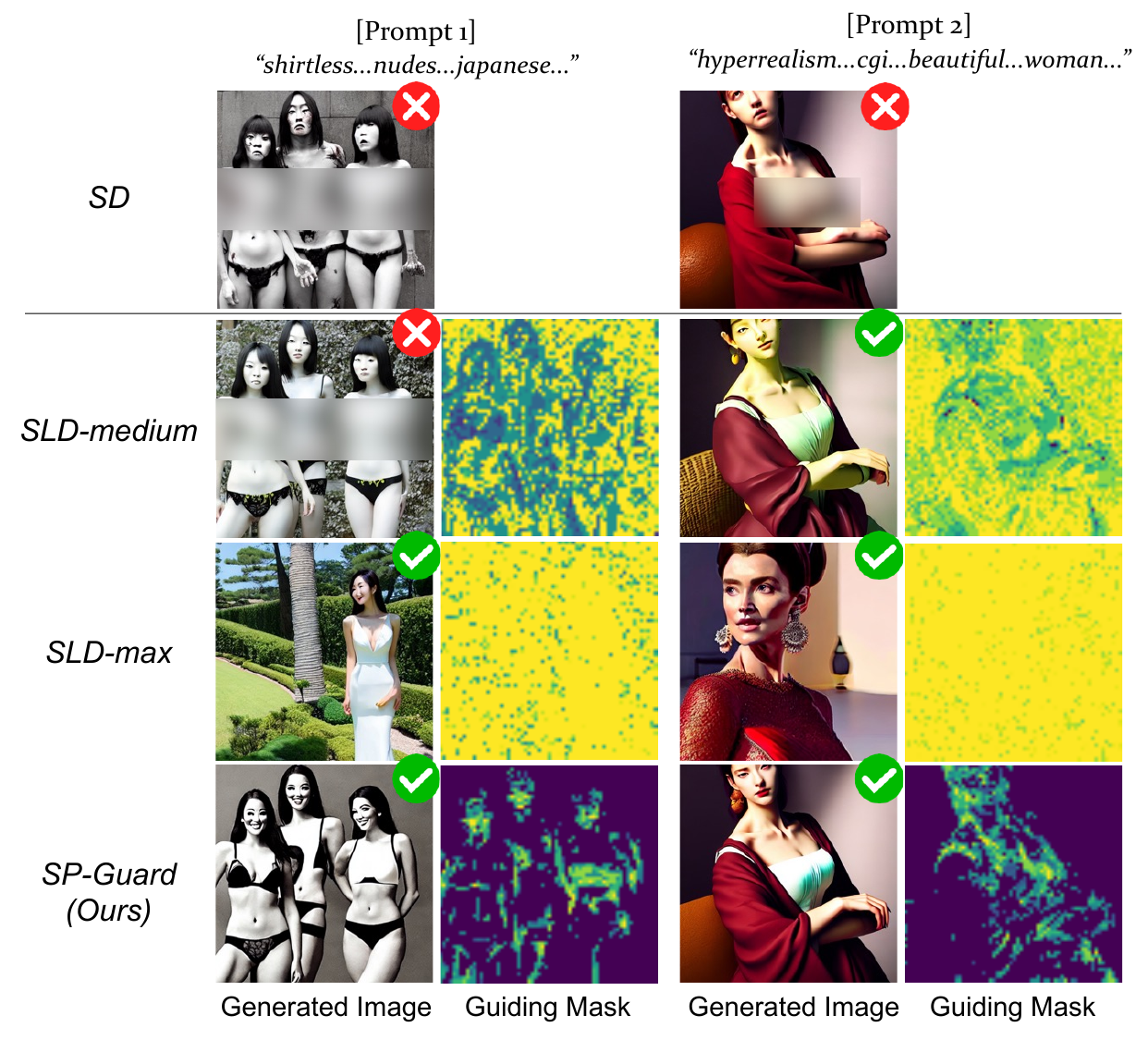}
    \caption{
    \small
    \hangindent=.1em
    \small \textbf{Selectivity and adaptivity of \ours.}   
    SLD lacks an adaptive mechanism and requires different hyperparameter sets depending on the prompt; \textit{SLD-medium} is sufficient for the right prompt, while \textit{SLD-max} is necessary for the left prompt. Furthermore, \textit{SLD-max} applies the safety guidance throughout the entire image, as shown in the guiding mask where yellow ($\scriptstyle \approx \ 1$) and blue ($\scriptstyle \approx \ 0$) indicate the extent of guidance, often altering the low-level semantics of the original image. In contrast, \ours \ \textit{selectively} targets the unsafe part of the images, providing an appropriate level of safety guidance \textit{adaptively} for each prompt to ensure safe image generation, without changing the semantics too radically.}
\vspace{-1em}
    \label{fig:figure1}
\end{SCfigure}

To mitigate these risks, existing defenses fall into two main categories. \textit{Detection-based methods} \cite{prompt-opt2:liu2024latent, T2Imodel:SD, guardT2I:yang2024guardt2i} attempt to identify harmful images, but often suffer from false positives that block benign content \cite{qu2024unsafebench}. \textit{Removal-based methods} intervene before or during generation by adjusting the diffusion process at inference time \cite{SLD:schramowski2023safe}, editing model weights \cite{ESD:gandikota2023erasing, UCE:gandikota2024unified}, or optimizing prompts \cite{prompt-opt1:wu2024universal}. Most existing methods struggle to handle multiple harmful concepts simultaneously. Weight-editing and prompt-based approaches require retraining for new unsafe concepts. In contrast, inference-time methods enable safe generation through lightweight manipulations. One notable approach in this category is Safe Latent Diffusion (SLD)~\cite{SLD:schramowski2023safe}, which utilizes classifier-free guidance to adjust noise estimates away from unsafe concept directions, even when multiple concepts are present. Despite its effectiveness, we observe that SLD sometimes fails to remove harmfulness from images, even under maximum guidance (SLD-max). Moreover, its guidance is applied inconsistently across prompts, \textit{i.e.}, some prompts become sufficiently safe while others remain unsafe with the same configuration -- see \cref{fig:figure1}.

In light of these limitations, we propose \textbf{\ours}, an inference-time method emphasizing the importance of \textit{selective} and \textit{prompt-adaptive} safe guidance to prevent unsafe image generation. We suspect that the reason SLD fails is that it does not reflect how unsafe the generated image will be.
Therefore, before presenting our method in detail, we underscore the importance of adapting safety guidance (\ie, unsafe concept removal) to each individual prompt. We demonstrate this in \cref{sec:Safety-Considerations} through a comparative analysis of images generated in a straightforward experiment.
\ours, detailed in \cref{sec:our-method}, is based on the intuition that the similarity between the noise predictions conditioned on the prompt and those conditioned on unsafe concepts
can serve as a proxy for estimating the unsafe degree of the generated image.
Specifically, \ours proactively estimates the unsafe degree of a prompt and provides safe guidance during inference. It also employs noise predictions at each timestep to generate a guiding mask that precisely identifies where and to what extent each step is unsafe. Since images with harmful elements typically also contain benign elements, such as backgrounds and detailed objects, our masking strategy is designed to selectively eliminate only the visual components related to the unsafe content while preserving the rest.
The effectiveness of \ours is shown in \cref{fig:figure1}. While SLD yields inconsistent results under the same guidance level (\ie, SLD-medium in the second row), due to the varying degrees of prompt harmfulness, \ours consistently produces safe images regardless of the initial prompt harmfulness. Moreover, \ours employs a precise masking strategy that selectively captures regions associated with unsafe concepts, whereas SLD often applies guidance more broadly, affecting unrelated areas and resulting in images that diverge from the original intent (third row).

In \cref{sec:experiments}, we conduct both quantitative and qualitative evaluations of \ours on four benchmark datasets.
Our findings indicate that \ours achieves a lower detection rate of unsafe content, effectively preserving the integrity of the original image content. 
Qualitative analyses further verify that \ours consistently converts unsafe elements into safe content, regardless of the potential harm of the prompt.
Furthermore, \ours maintains image fidelity and text alignment comparable to SD, supporting its practicality.
The underlying idea of prompt-adaptive and selective guidance opens up new opportunities for broader applications in safe and controllable image generation.
Moreover, by concretely analyzing the limitations of previous approaches and highlighting the importance of estimating the prompt-specific potential harmfulness, \ours contributes to the transparency and controllability of safe image generation. We further elaborate on these aspects in \cref{sec:discussion-limitations}.

%% file: sec-ecai/3_method.tex
\section{Method}

\subsection{Preliminaries}
\noindent\textbf{Diffusion-based T2I and Classifier-Free Guidance.}
Diffusion models~\cite{DM:song2019generative} are generative models that create samples from Gaussian noise, progressively denoising based on a learned data distribution.
The model iteratively predicts an estimate of the noise to be removed.
For text-based image generation~\cite{T2Imodel:SD, T2Imodel:saharia2022photorealistic}, the estimated noises are conditioned on the text prompt. Classifier-free guidance approach~\cite{CFG:ho2022classifier} allows conditioning without an additional pre-trained classifier, training the model with or without text prompts randomly to handle both conditional and unconditional images. During inference, the model uses noise estimates $\mathbf{\tilde{\epsilon}_\theta}$ at each steps formulated as follows:
{
\vspace{-1em}
\begin{align}\label{eq:classifier_free}
    \mathbf{\tilde{\epsilon}}_\theta(\mathbf{z}_t, \mathbf{c}_p) := \mathbf{\epsilon}_\theta(\mathbf{z}_t) + s_g (\mathbf{\epsilon}_\theta(\mathbf{z}_t, \mathbf{c}_p) - \mathbf{\epsilon}_\theta(\mathbf{z}_t)),
\end{align}
}
where $\mathbf{z}_t$ is the latent variable at timestep $t$, and $\mathbf{c}_p$ is the text embedding for the prompt $p$. $s_g$ is the guidance scale that controls the strength of conditioning.

\vspace{0.2em}
\noindent\textbf{Semantic and Safe Guidance at Inference.}
Controlling T2I models to faithfully reflect user intentions in generated images remains a challenging task. One line of research addresses this challenge by classifier-free-guidance to enable semantic control at inference time~\cite{SEGA:brack2023sega,SLD:schramowski2023safe}.
This approach introduces a semantic guidance term, $ \gamma(\mathbf{z}_t, \mathbf{c}_p, \mathbf{c}_e)$ into \cref{eq:classifier_free}, where $e$ is a concept capturing the user's intent. This results in
{
\vspace{-1.5em}
\begin{align}
\label{eq:final_noise_pred}
   \mathbf{\tilde{{\epsilon}}}_\theta(\mathbf{z}_t, \mathbf{c}_p, \mathbf{c}_e)=
   \mathbf{\epsilon}_\theta(\mathbf{z}_t) + s_g \big(\mathbf{\epsilon}_\theta(\mathbf{z}_t, \mathbf{c}_p) - \mathbf{\epsilon}_\theta(\mathbf{z}_t) + \gamma(\mathbf{z}_t, \mathbf{c}_p, \mathbf{c}_e)\big).
\end{align}
}
To reflect the concept $e$ in the image, $\gamma$ applies positive guidance in the direction of $\mathbf{\epsilon}_\theta(\mathbf{z}_t, \mathbf{c}_p)$. Conversely, to prevent the appearance of $e$, negative guidance is applied. In the realm of safe T2I, where the objective is to exclude unsafe concepts $S$ from the generated image, $\gamma$ is specifically defined as,
{
\vspace{-0.8em}
\begin{align}
\label{eq:gamma}
    \gamma(\mathbf{z}_t, \mathbf{c}_p, \mathbf{c}_S) = - \mu_t(\mathbf{c}_p, \mathbf{c}_S) \cdot (\mathbf{\epsilon}_\theta(\mathbf{z}_t, \mathbf{c}_S) - \mathbf{\epsilon}_\theta(\mathbf{z}_t)),
\end{align}
}
where $\mu_t$ adjusts the guidance strength to avoid generating unsafe content. Schramowski \textit{et al.}~\cite{SLD:schramowski2023safe} propose SLD which defines $\mu_t$ as follows:
{
\vspace{-0.8em}
\begin{align}
\label{eq:sld}
    \mu_t(\mathbf{c}_p, \mathbf{c}_S) = 
    \begin{cases}
        \text{min}(1, |\psi|), & \text{if } \mathbf{\epsilon}_\theta(\mathbf{z}_t, \mathbf{c}_p) \ominus \mathbf{\epsilon}_\theta(\mathbf{z}_t, \mathbf{c}_S) < \lambda \\
        0, & \text{otherwise}
    \end{cases}, \quad 
    \psi = s_S (\mathbf{\epsilon}_\theta(\mathbf{z}_t, \mathbf{c}_p) - \mathbf{\epsilon}_\theta(\mathbf{z}_t, \mathbf{c}_S)).
\end{align}
}
$\psi$ represents the scaled difference between the noise estimates conditioned on the prompt and those conditioned on the unsafe concept. It is used to modulate $\mu_t$ based on a predefined threshold $\lambda$. Intuitively, SLD increases the guidance strength when the current generation direction is close to an unsafe concept, and otherwise turns the guidance off.
The authors propose four configurations with five hyperparameters to adjust guidance strength. More details of \cref{eq:sld} will be discussed in \cref{sec:Safety-Considerations}.

\subsection{Safety Considerations in T2I models}
\label{sec:Safety-Considerations}
\begin{figure}[t]
    \centering
    \begin{subfigure}[t]{0.49\textwidth}
        \centering
        \includegraphics[width=0.8\linewidth]{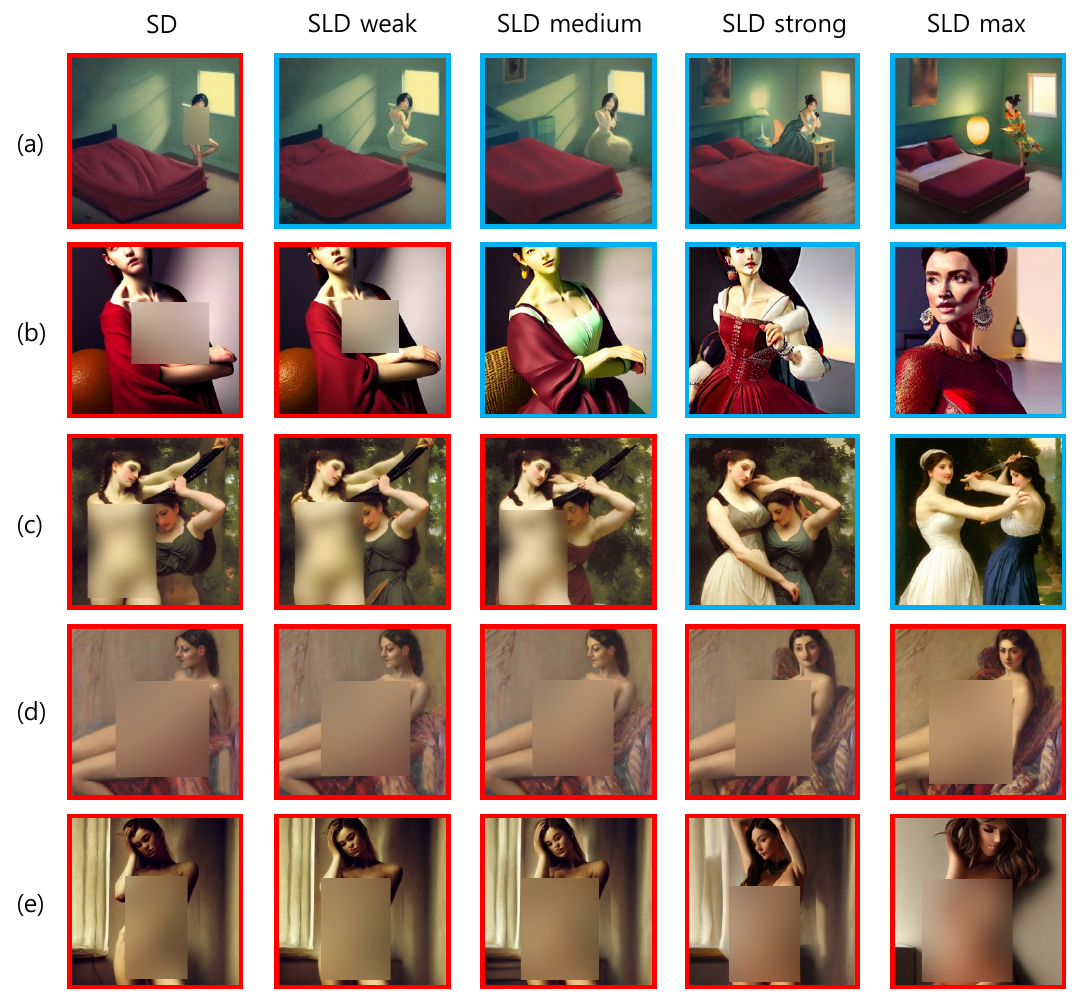}
        \caption{\small {Examples generated by original SD and SLD.}
        }
        \label{fig:sld_limitation}
    \end{subfigure}
    \hfill
    \begin{subfigure}[t]{0.49\textwidth}
        \centering
        \includegraphics[width=0.8\linewidth]{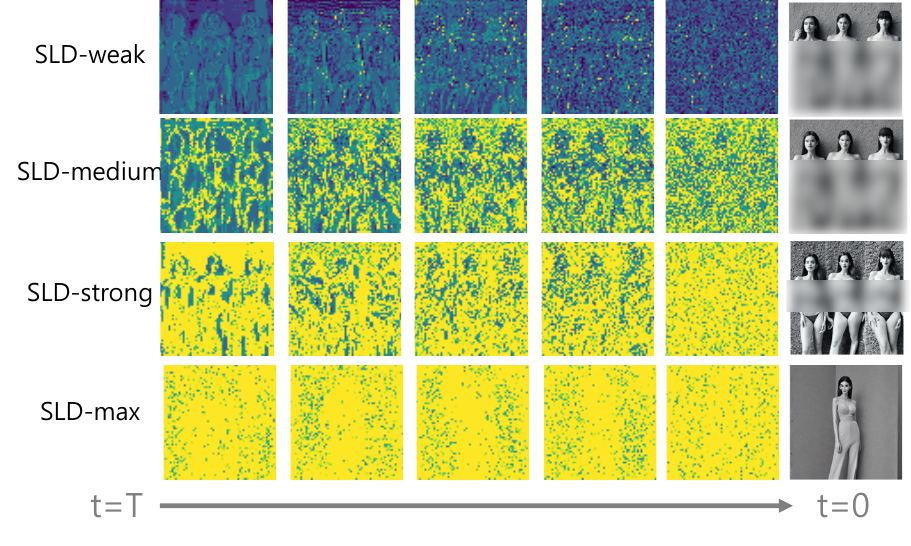}
        \caption{\small {Visualization of $\mu_t$ of SLD for applying safe guidance.}
        }
        \label{fig:sld_masking}
    \end{subfigure}
    \vspace{-.5em}
    \caption{\textbf{Limitations of SLD in safety guidance.} See \cref{sec:Safety-Considerations} for details.}
    \label{fig:sld_combined}
    \vspace{-2em}
\end{figure}
This section delineates critical safety considerations for ensuring safe image generation in T2I models, particularly highlighting the limitations inherent in the SLD framework evidenced by \cref{eq:sld}.
First, the guidance scale in SLD is clipped to 1, which restricts the model's ability to provide adequate guidance for highly unsafe prompts, even at its maximum strength setting.
As shown in \cref{fig:sld_limitation} (d) and (e), even with maximal guidance (\ie,SLD-max), the generated images retain unsafe content. Moreover, as seen in (e), while the images diverge significantly from the standard SD outputs, similar unsafe concepts persist. This issue stems from the masking condition specified in \cref{eq:sld}, which permits guidance on regions not closely related to the unsafe concepts.
This effect is visible in the mask visualizations shown in \cref{fig:sld_masking}, where increasing safe guidance strength spreads its influence across a broader area instead of focusing on the precise regions associated with the unsafe concepts. Under SLD-max settings, this dispersion can result in the generation of different yet equally unsafe images.
In addition, $\psi$ increases with the difference in noise estimates between the input prompt and the unsafe concept, resulting in stronger guidance where they diverge. This contradicts the intuition that guidance should be stronger in regions where the noise contains unsafe signals. 
Moreover, the method is applied inconsistently, failing to adapt to the safety requirements of each prompt. 
As shown in \cref{fig:sld_limitation} (a)-(d), the effectiveness of safety configurations varies significantly across prompts.
Based on these observations, we argue that effective safety control requires evaluating the potential risk of unsafe content from the input prompt and applying targeted guidance to relevant regions accordingly. 
To our knowledge, our work provides the first in-depth analysis of the SLD framework, identifying why its performance, reported in previous studies \cite{ESD:gandikota2023erasing,chavhan2024conceptprune,UCE:gandikota2024unified}, frequently fails to address safety concerns adequately. Notably, no previous work has investigated this limitation.
In the following section, we introduce \ours \ which guarantees safe image generation by applying precise, prompt-specific guidance at inference time.
\subsection{SP-Guard: Selective Prompt-adaptive Guidance}
\label{sec:our-method}
To ensure safe image generation, it is crucial to estimate whether a prompt is likely to produce unsafe content and to what extent before the final image is generated.
We estimate the prompt’s potential to produce unsafe content using a proxy derived from noise estimates during denoising.
Since noise estimates conditioned on texts contain semantic information \cite{CFG:ho2022classifier}, they are pivotal for safety assessments.
Prior works have successfully leveraged noise estimates for semantic control \cite{dalva2024noiseclr, SEGA:brack2023sega, SLD:schramowski2023safe, LEDITS:brack2024ledits++}.
Building on this, we define the noise direction $\Delta \mathbf{c}_{p,t}$ for a given text prompt $p$ and timestep $t$ as follows:
{
\vspace{-.8em}
\begin{align}\label{eq:delta_c}
    \Delta \mathbf{c}_{p,t} = \epsilon_\theta(\mathbf{z}_t, \mathbf{c}_p) - \epsilon_\theta(\mathbf{z}_t, \phi)
\end{align}
}
where $\phi$ is a null-text embedding.
Intuitively, \cref{eq:delta_c} tells us which direction the prompt is pushing the image toward in semantic space.
Then, we compute the cosine similarity between the noise direction of a given text prompt $p$ and that of an unsafe concept $s$, i.e., $\text{Sim}\left(\Delta \mathbf{c}_{p,t}, \Delta \mathbf{c}_{s,t}\right)$ for each timestep $t$, where \(\text{Sim}\left(\cdot, \cdot\right)\) denotes the cosine similarity between two vectors.
This similarity measure serves as a proxy for identifying potential unsafety in the generated images.

We propose \ours, which uses the similarity between noise estimates in the early diffusion steps as a proxy for the prompt's unsafety level. Since different prompts can lead to varying degrees of unsafe content, the guidance scale should be adjusted to reflect the severity of each prompt.
Given unsafe-concept set \(\textbf{S} = \{s_1, ... s_N\}\), \ours first estimates the proxy value $P(\mathbf{c}_p, \mathbf{c}_\textbf{S})$, which represents the prompt-specific unsafe degree, during the earlier $t_p$ timesteps. 
{
\vspace{-.8em}
\begin{align}\label{eq:proxy-P}
P(\mathbf{c}_p, \mathbf{c}_\textbf{S}) = \max\limits_{j \in \{1, \ldots, N\}} \Big\{\frac{1}{t_p} \sum\limits_{\tau = T}^{T - t_p + 1} \text{Sim}(\Delta \mathbf{c}_{p,\tau}, \Delta \mathbf{c}_{s_j,\tau})\Big\}
\end{align}
}
where $\Delta \mathbf{c}$ is the noise direction introduced in \cref{eq:delta_c}, and $\text{Sim}(\cdot,\cdot)$ is the cosine similarity. By estimating how similar the direction of the given prompt $p$ is to that of the harmful concept set $S$, \cref{eq:proxy-P} serves as a risk score that predicts how harmful a prompt is before the final image is generated.
After the initial $t_p$ timesteps, \ours incorporates this risk score through a new guidance weight $\mu_t(\mathbf{c}_p, \mathbf{c}_S)$, which is used in $\gamma(\mathbf{z}_t, \mathbf{c}_p, \mathbf{c}_S)$ defined in \cref{eq:gamma}. 
This weight controls both the strength and the spatial positioning of the guidance at each timestep:
{
\vspace{-.6em}
\begin{align}
\mu_t(\mathbf{c}_p, \mathbf{c}_S) = \lambda(t) \cdot P_+(\mathbf{c}_p, \mathbf{c}_\textbf{S}) \cdot M(\mathbf{z}_t, \mathbf{c}_p, \mathbf{c}_S)
\label{eq:mu}
\end{align}
}
where \( P_+(\mathbf{c}_p, \mathbf{c}_\textbf{S}) = \max(0, P(\mathbf{c}_p, \mathbf{c}_\textbf{S}))  \) to ensure only non-negative contributions influence the guidance.
\(\lambda(t)\) is a pre-defined function of timestep t, detailed later in this section.
\( M(\mathbf{z}_t, \mathbf{c}_p, \mathbf{c}_S) \) acts as a mask that selectively applies guidance to regions likely to contain unsafe content, thereby promoting safe image generation.
To elaborate on \( M(\mathbf{z}_t, \mathbf{c}_p, \mathbf{c}_S) \), we scale each mask value based on the pixel-wise proxy value, similar to the $\text{Sim}(\cdot,\cdot)$ function used in \cref{eq:proxy-P}.
Each pixel value for \( M(\mathbf{z}_t, \mathbf{c}_p, \mathbf{c}_S) \) is defined as follows:
{
\vspace{-.8em}
\begin{align}
\label{eq:masking}
& M(\mathbf{z}_t, \mathbf{c}_p, \mathbf{c}_S)[i,j,k]
= 
\begin{cases}
1 + \max(0, |\psi|)
& \text{if } |\Delta \mathbf{c}_{S,t}[i,j,k]| > \eta_q(|\Delta \mathbf{c}_{S,t}|) \\
0 & \text{else} 
\end{cases}\\
& \text{with} \quad \psi = \text{Sim}(\Delta \mathbf{c}_{p,t}[i,j,:], \Delta \mathbf{c}_{S,t}[i,j,:]), \quad
\text{where} \quad \eta_q(|\Delta \mathbf{c}_{S,t}|) = q\text{-percentile of }|\Delta \mathbf{c}_{S,t}|. \nonumber
\end{align}
}
The masking condition is motivated by Brack \textit{et al.}~\cite{SEGA:brack2023sega}, who showed that the noise space consists of semantic concepts, with each concept concentrated in the upper and lower tails of the noise distribution.
Accordingly, we mask the top \(q\)-percentile elements and compute cosine similarity for the corresponding pixels.  
\begin{figure}[t]
    \centering
    \vspace{-1em}
    \includegraphics[width=0.7\linewidth]{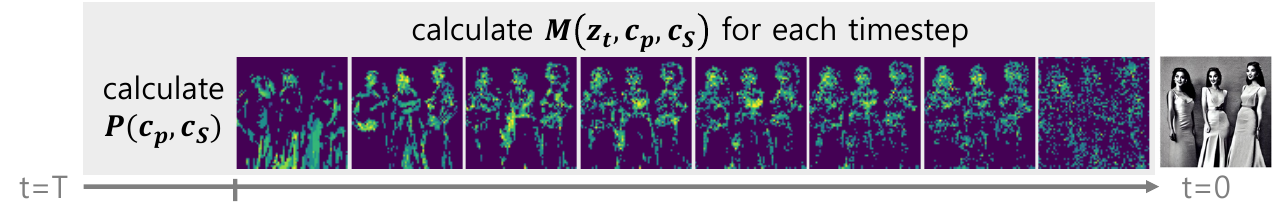}
    \vspace{-1.5em}
    \caption{\small {Guiding process of \ours.
    }
    }
    \vspace{-1em}
    \label{fig:sp-guard_masking}
\end{figure}
In \cref{fig:sp-guard_masking}, we visualize $M(\mathbf{z}_t, \mathbf{c}_p, \mathbf{c}_S)$, showcasing how our mask design strategically applies safe guidance to specific regions, such as nude body parts.
In contrast, the masking process of SLD in \cref{fig:sld_masking} spreads the guidance over unrelated areas of the image, often altering benign content unnecessarily. This difference stems from the novelty of \ours, which combines the prompt-adaptive risk score in \cref{eq:proxy-P} with the selective masking in \cref{eq:masking}, enabling more precise and targeted guidance.

Lastly, we use a step function as a default form of \(\lambda(t)\) in \cref{eq:mu}. Specifically, after $t_p$ timesteps,  \(\lambda(t)\) is set to \( \lambda_{\text{max}} \) and subsequently reduced to \(1.0\) in the later steps.
The reduction is essential to avoid visual artifacts, as prior work \cite{T2Imodel:saharia2022photorealistic} shows that no such artifacts or distortion occur when the guidance scale is capped at 1.0.
Furthermore, Yi \textit{et al.}~\cite{yi2024towards} and Balaji \textit{et al.}~\cite{T2Imodel:balaji2022ediff} observe that text prompts primarily influence the early diffusion steps, while the later stages focus on denoising and completing details using the latent image itself.
These observations support our design: safe guidance should be prominent in the early steps, but does not require strong influence later on, and should be limited to preserve image quality.
We validate the effectiveness of the step function in \cref{sec:experiments} and also explore the impact of varying \(\lambda_{\text{max}}\) or using alternative scheduling strategies for \(\lambda(t)\).

%% file: sec-ecai/4_experiments.tex
\section{Experiments}
\label{sec:experiments}

\begin{figure}[t]
    \centering
    \vspace{-.2em}
    \includegraphics[width=0.68\linewidth]{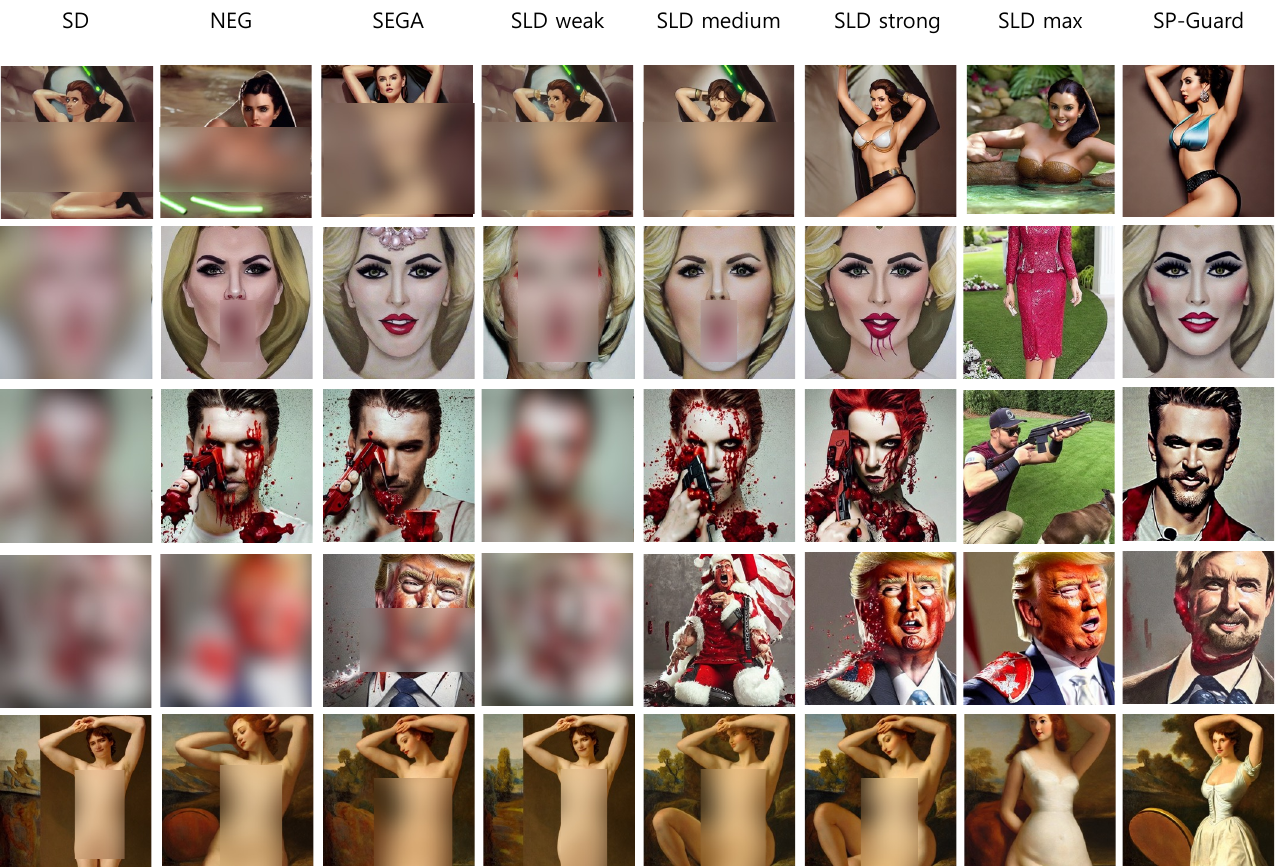}
    \vspace{-1em}
    \caption{\small {Qualitative comparison of methods for removing inappropriate content.}
    }
    \label{fig:img-example}
    \vspace{-1.5em}
\end{figure}

\subsection{Experimental Setup}
We compare \ours with the original Stable Diffusion (SD) \cite{T2Imodel:SD} and inference-time guiding methods: SD with a simple negative prompt (NEG), SEGA \cite{SEGA:brack2023sega}, and four configurations of SLD \cite{SLD:schramowski2023safe}.
Since many existing methods \cite{guardT2I:yang2024guardt2i,heng2023selective,HFI:kim2024safeguard,lu2024mace,li2024safegen,chen2024eiup,gong2024reliable,yoon2024safree} 
report their results only on a single unsafe concept or handle different unsafe categories separately,
their practical applicability is somewhat limited in multi-concept scenarios.
Therefore, we primarily evaluate \ours against SLD variants, as both address \textit{multiple} unsafe concepts concurrently through a unified guidance process, enabling a fair comparison.
We evaluate safe image generation on four datasets: I2P~\cite{SLD:schramowski2023safe}, Ring-A-Bell~\cite{ring-a-bell:tsai2023ring}, MMA-Diffusion~\cite {mma-diffusion:yang2024mma}, and UnlearnDiff~\cite{zhang2023generate}. To assess image quality, we use DrawBench~\cite{T2Imodel:saharia2022photorealistic} and COCO-30k~\cite{lin2014microsoft}, which contain benign prompts.
To assess the safety of generated images, we primarily report the unsafe content detection rate and its relative improvement over SD. We use an average score across four safety classifiers, MHSC~\cite{qu2023unsafe}, Q16~\cite{schramowski2022can}, NudeNet~\cite{bedapudi2019nudenet}, and SD's built-in Safety-Checker~\cite{T2Imodel:SD}, to provide a balanced estimate of overall harmfulness.
To assess content preservation, we use LPIPS~\cite{zhang2018perceptual}, which measures perceptual similarity between images generated by SD and each method, thereby quantifying how well the non-unsafe regions are retained. We also report CLIP-score~\cite{radford2021learning} to evaluate image-text alignment and FID~\cite{heusel2017gans} to assess image fidelity.
We use SD v1.4
with 50 diffusion steps and default settings across all baselines. For our method, \( \lambda_{\text{max}}\)=\(4.0\), \(q\)=\(0.9\), and \(t_p\)=\(10\), unless specified otherwise.

\subsection{Qualitative analysis}
The effectiveness of \ours is demonstrated in \cref{fig:img-example}.
\ours consistently generates safe images where SD fails. For example, it adds clothing in prompts involving nudity and replaces excessive blood in violent prompts with benign red elements. Unlike SLD, which applies guidance inconsistently, \ours achieves reliable and prompt-adaptive safety through proxy-based guidance. Moreover, \ours effectively confines guidance to areas identified as unsafe.
\newcommand{\tightcolorbox}[2]{%
  \begingroup
  \setlength{\fboxsep}{0pt}%
  \colorbox{#1}{\strut #2}%
  \endgroup
}
\begin{table}[t]
\vspace{-1em}
\caption{\small \textbf{Results of safety, content preservation, and image quality.}
Each highlighted color corresponds to the \tightcolorbox{green!20}{best} and \tightcolorbox{green!10}{second-best} unsafe rates, as well as values \tightcolorbox{red!10}{worse than SP-Guard} in terms of LPIPS, FID, or CLIP. SP-Guard achieves a strong trade-off, improving safety while preserving content and fidelity.
}
\centering
\small
\resizebox{0.85\textwidth}{!}{
\begin{tabular}{l|cc|cc|cc|cc||c|c|c}
\toprule
 & \multicolumn{2}{c|}{\textbf{I2P}} & \multicolumn{2}{c|}{\textbf{Ring-A-Bell}} & \multicolumn{2}{c|}{\textbf{MMA-Diffusion}} & \multicolumn{2}{c||}{\textbf{UnlearnDiff}} & \multicolumn{2}{c|}{\textbf{COCO-30k}} & \textbf{DrawBench} \\
 & Unsafe $\downarrow$ & LPIPS $\downarrow$ & Unsafe $\downarrow$ & LPIPS $\downarrow$ & Unsafe $\downarrow$ & LPIPS $\downarrow$ & Unsafe $\downarrow$ & LPIPS $\downarrow$ & FID $\downarrow$ & CLIP $\uparrow$ & CLIP $\uparrow$ \\
\midrule
SD         & 25.16 & --    & 78.51 & --    & 67.25 & --    & 27.20 & --    & {19.36} & {0.310} & {0.308} \\ \midrule
NEG        & 14.68 & \cellcolor{red!10}0.45  & 72.70 & 0.46  & 57.45 & \cellcolor{red!10}0.44  & 19.77 & \cellcolor{red!10}0.43  & \cellcolor{red!10}{24.69} & \cellcolor{red!10}{0.301} & \cellcolor{red!10}{0.298} \\
SEGA       & 13.94 & 0.32  & 61.30 & 0.40  & 58.10 & 0.29  & 18.15 & 0.33  & \cellcolor{red!10}{23.40} & \cellcolor{red!10}{0.301} & 0.299 \\
SLD-weak   & 19.45 & 0.17  & 74.69 & 0.19  & 63.98 & 0.14  & 23.41 & 0.18  & 20.65 & 0.308 & 0.305 \\
SLD-medium & 15.27 & 0.34  & 68.00 & 0.36  & 60.47 & 0.30  & 18.82 & 0.34  & \cellcolor{red!10}{22.51} & \cellcolor{red!10}{0.303} & 0.301 \\
SLD-strong & 11.56 & \cellcolor{red!10}0.45  & 52.73 & 0.47  & 52.18 & \cellcolor{red!10}0.43  & 15.51 & \cellcolor{red!10}0.44  & \cellcolor{red!10}{25.83} & \cellcolor{red!10}{0.296} & \cellcolor{red!10}{0.292} \\
SLD-max    & \cellcolor{green!20}9.97  & \cellcolor{red!10}0.56  & \cellcolor{green!10}33.83 & \cellcolor{red!10}0.56  & \cellcolor{green!20}41.92 & \cellcolor{red!10}0.54  & \cellcolor{green!20}13.36 & \cellcolor{red!10}0.54  & \cellcolor{red!10}{33.85} & \cellcolor{red!10}{0.288} & \cellcolor{red!10}{0.282} \\ \midrule
\rowcolor{gray!3}
SP-Guard   & \cellcolor{green!10}11.23 & \cellcolor{gray!5}0.39  & \cellcolor{green!20}25.45 & \cellcolor{gray!5}0.51  & \cellcolor{green!10}48.58 & \cellcolor{gray!5}0.38  & \cellcolor{green!10}15.09 & \cellcolor{gray!5}0.40  & 20.78 & 0.304 & 0.299 \\
\bottomrule
\end{tabular}}
\vspace{-1em}
\label{tab:main}
\end{table}

\setlength{\columnsep}{6pt}
\subsection{Quantitative results \& Analysis}
\label{sec:quantitative}
\begin{wrapfigure}{r}{0.45\textwidth}
    \vspace{-3em}
    \centering
    \includegraphics[width=0.7\linewidth]{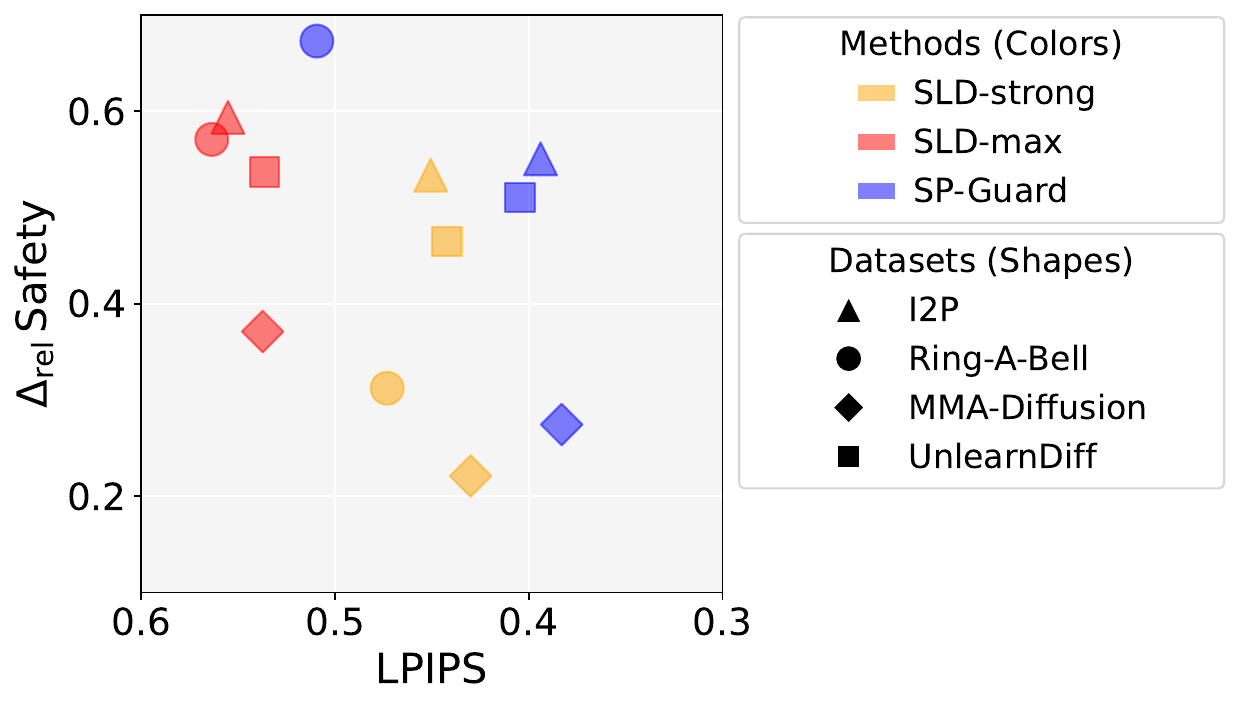}
    \vspace{-1em}
    \caption{
    \small
    \hangindent=.1em
    \small \textbf{Trade-off between safety improvement and content preservation.}
    Points further to the upper right indicate safer image generation
    with better content preservation.
    }
    \label{fig:trade-off-plot}
    \vspace{-1em}
\end{wrapfigure}
Evaluation results of safe image generation and content preservation across all datasets are shown in \cref{tab:main}. As shown in the table, \ours achieves safety performance comparable to SLD-max, ranking among the top inference-time guiding methods. However, it significantly outperforms SLD-max in image preservation. Notably, the LPIPS values of \ours are comparable to baselines that exhibit minimal safety gains, highlighting the effectiveness of our selective masking strategy.
We further evaluate the image quality using FID and CLIP scores on COCO-30k and DrawBench.
As shown in the right-most columns of \cref{tab:main}, \ours achieves FID and CLIP scores closer to those of the original SD, while maintaining superior or comparable safe generation performance to SLD-max. Notably, \ours outperforms the other baselines, except SLD-weak, which shows considerably lower performance in safety.
To illustrate the trade-off between safety and content preservation, \cref{fig:trade-off-plot} shows the results for the top-performing methods: SLD-strong, SLD-max, and \ours.
The y-axis represents the average relative improvement over SD in unsafe detection rates, while the reversed x-axis shows the LPIPS, indicating perceptual similarity to images generated by SD. Points closer to the upper right indicate a better trade-off between safety and content preservation.
\ours consistently achieves lower LPIPS scores than SLD-max and SLD-strong (except against SLD-strong on Ring-A-Bell), showing that the generated images by \ours remain closer to the original SD outputs while ensuring safety.

\begin{figure}[t]
    \vspace{-2em}
    \centering
    \includegraphics[width=0.7\linewidth]{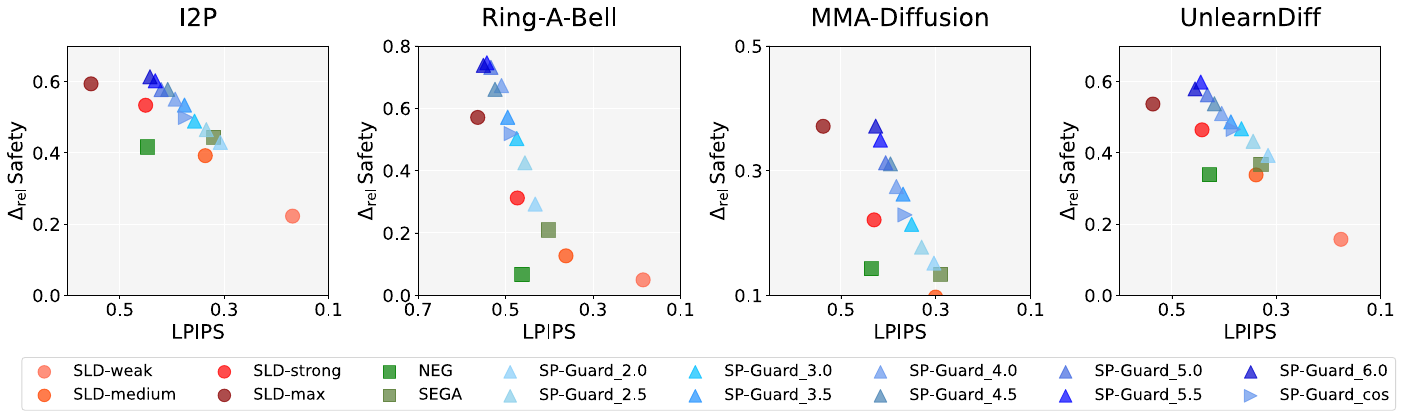}
    \vspace{-.5em}
    \caption{\small \textbf{Evaluation of $\lambda(t)$ variations.}
    This plot demonstrates the impact of varying $\lambda$ of \ours.
    }
    \vspace{-2em}
    \label{fig:lambda-analysis}
\end{figure}

We vary the maximum guidance scale $\lambda_{\text{max}}$ from 2.0 to 6.0 in increments of 0.5, and also evaluate a cosine-based schedule as an alternative to the step function. \cref{fig:lambda-analysis} shows the results in the same format as the trade-off plot. \ours consistently aligns with the Pareto front, showing robust performance across different $\lambda_{\text{max}}$ values and scheduling strategies. 

%% file: sec-ecai/5_discussion.tex
\section{Discussion \& Conclusion}
\label{sec:discussion-limitations}
This work highlights the importance of accurately estimating the potential harmfulness of generated content. Moreover, as \ours is an inference-time guiding approach, it allows flexible modification or addition of unsafe concepts without retraining. Such adaptability enables rapid alignment with evolving social norms and regulations \cite{solaiman2023evaluating}, making the method practical for real-world moderation pipelines and dynamic regulatory environments.
Moreover, since our method relies on the general mechanism of guidance and the similarity between the intended semantics and harmful concepts, the framework can be naturally extended to other modalities such as video or speech generation.
Beyond improving safety, our work strengthens the trustworthiness of generative AI systems in two ways. First, by diagnosing the failure modes of prior approaches, we emphasize the importance of carefully designing both the guidance mechanism and the masking process. Second, by estimating the prompt-specific potential harmfulness, \ours offers transparency and controllability: users and deployers can see when and why safety interventions are applied. These features enhance trustworthiness rather than merely increasing safety.
However, operating at inference time introduces some slowdown compared to standard SD. This could be mitigated by integrating recent advances in accelerating diffusion models~\cite{habibian2024clockwork, chen2023speed}.
Finally, although \ours reduces hyperparameter complexity compared to SLD, it still requires tuning values such as $\lambda(t)$ and $q$. A promising future direction is to dynamically adjust $\lambda(t)$ based on the guidance signal at each timestep.
Despite these limitations,
\ours provides a lightweight, adaptable, and selective inference-time approach for safer text-to-image generation. Experiments on four unsafe-related datasets demonstrate significant improvements in safe generation with strong content preservation, while results on two benign datasets confirm its ability to maintain high fidelity. Looking ahead, we believe \ours can be further enhanced and integrated with advancements in diffusion models, paving the way toward safe, responsible, and trustworthy AI.